\documentclass{article}

\PassOptionsToPackage{numbers, compress}{natbib}
\usepackage[preprint]{neurips_2026}

\usepackage[utf8]{inputenc} 
\usepackage[T1]{fontenc}    
\usepackage{hyperref}       
\usepackage{url}            
\usepackage{booktabs}       
\usepackage{amsfonts}       
\usepackage{nicefrac}       
\usepackage{microtype}      

\usepackage{graphicx}       
\usepackage{soul, color}    
\usepackage[table]{xcolor}  
\usepackage{caption}

\usepackage{enumitem}
\usepackage{xfrac}
\usepackage{makecell}
\usepackage{multirow}
\usepackage{amsmath}
\usepackage[normalem]{ulem}
\usepackage{rotating}
\usepackage{pifont}

\usepackage{graphicx}      
\usepackage{booktabs}      
\usepackage{rotating}      
\usepackage[table]{xcolor} 
\usepackage{amssymb}       
\usepackage{pifont}        
\usepackage{multirow}      

\usepackage{wrapfig} 

\newcommand{\first}[1]{\textbf{#1}}
\newcommand{\second}[1]{\underline{#1}}

\definecolor{barriercolor}{RGB}{255, 120, 50}
\definecolor{bicyclecolor}{RGB}{255, 192, 203}
\definecolor{buscolor}{RGB}{255, 255, 0}
\definecolor{carcolor}{RGB}{0, 150, 245}
\definecolor{constructcolor}{RGB}{0, 255, 255}
\definecolor{motorcolor}{RGB}{200, 180, 0}
\definecolor{pedestriancolor}{RGB}{255, 0, 0}
\definecolor{trafficcolor}{RGB}{255, 240, 150}
\definecolor{trailercolor}{RGB}{135, 60, 0}
\definecolor{truckcolor}{RGB}{160, 32, 240}
\definecolor{drivablecolor}{RGB}{255, 0, 255}
\definecolor{otherflatcolor}{RGB}{139, 137, 137}
\definecolor{sidewalkcolor}{RGB}{75, 0, 75}
\definecolor{terraincolor}{RGB}{150, 240, 80}
\definecolor{manmadecolor}{RGB}{230, 230, 255}
\definecolor{vegetationcolor}{RGB}{0, 175, 0}
\definecolor{otherscolor}{RGB}{0, 0, 0}
\title{Height-Guided Projection Reparameterization for Camera-LiDAR Occupancy}

\author{%
  Yuan Wu\textsuperscript{\rm 1} \quad 
  Zhiqiang Yan\textsuperscript{\rm 2}\thanks{Corresponding authors.} \quad 
  Jiawei Lian\textsuperscript{\rm 1} \quad 
  Zhengxue Wang \textsuperscript{\rm 1} \quad 
  Jian Yang\textsuperscript{\rm 1}\footnotemark[1] \\
    \textsuperscript{\rm 1}Nanjing University of Science and Technology \\
    \textsuperscript{\rm 2}National University of Singapore  \quad
}

\begin{document}

\maketitle

\begin{abstract}
3D occupancy prediction aims to infer dense, voxel-wise scene semantics from sensor observations, where the 2D-to-3D view transformation serves as a crucial step in bridging image features and volumetric representations. 
Most previous methods rely on a fixed projection space, where 3D reference points are uniformly sampled along pillars. 
However, such sampling struggles to capture the sparsity and height variations of real-world scenes, leading to ambiguous correspondences and unreliable feature aggregation. 
To address these challenges, we propose \textbf{HiPR}, a camera-LiDAR occupancy framework with \textbf{H}eight-Gu\textbf{i}ded \textbf{P}rojection \textbf{R}eparameterization.
HiPR first encodes LiDAR into a BEV height map to capture the maximum height of the point cloud. 
HiPR then adjusts the sampling range of each pillar using the height prior, enabling adaptive reparameterization of the projection space. 
As a result, the projected points are redistributed into geometrically meaningful regions rather than fixed ranges. 
Meanwhile, we mask out the invalid parts of the height map to avoid misleading the feature aggregation. 
In addition, to alleviate the training instability caused by noisy LiDAR-derived heights, we introduce a training-time Progressive Height Conditioning strategy, which gradually transitions the conditioning signal from ground-truth heights to LiDAR heights.
Extensive experiments demonstrate that HiPR consistently outperforms existing state-of-the-art methods while maintaining real-time inference.
The code and pretrained models can be found at \url{https://github.com/yanzq95/HiPR}.
\end{abstract}

\section{Introduction}
3D occupancy prediction~\cite{ma2024cotr,wang2023openoccupancy,wei2023surroundocc,wu2025see,zhu2026dr} aims to estimate voxel-wise scene semantics from sensor observations and is crucial for holistic scene understanding and motion planning in autonomous driving~\cite{dang2026sparseworld,du2025sparseworld,liu2025occvla,yang2025driving,zheng2024occworld}. Camera-based occupancy prediction~\cite{cao2026occany,chen2025rethinking,chen2025semantic,leng2025occupancy,li2026ashsr} has attracted increasing attention, as cameras provide rich semantic cues at low cost and with wide coverage. To further improve performance, recent works~\cite{ming2024occfusion,wei2025ms,wolters2025unleashing,yang2025daocc,zheng2026doracamom} increasingly explore multi-modal frameworks that integrate complementary signals from LiDAR or radar.

\begin{figure}[t]
    \centering
    \includegraphics[width=1.0\columnwidth]{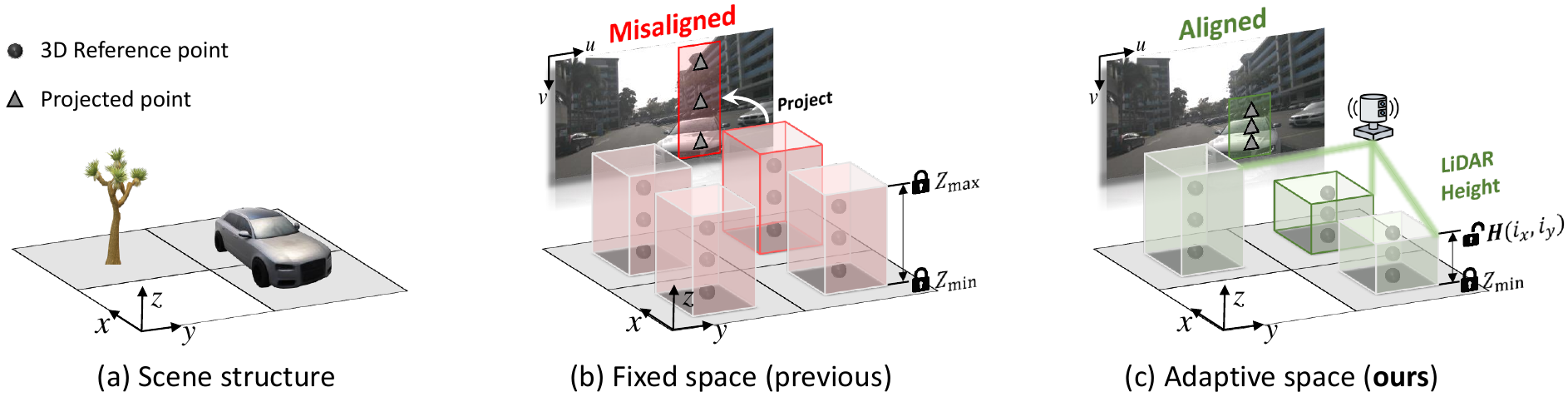}
   \caption{(a) Scene structure with varying heights. (b) Previous methods rely on a predefined fixed projection space, causing semantic misalignment on the 2D image plane. (c) Our method utilizes the LiDAR height prior to adaptively bound the sampling space of each pillar, ensuring alignment between projected reference points and scene structures for reliable feature aggregation.}
    \label{Fig:teaser}
    \vspace{-4pt}
\end{figure}

Despite these advances, accurately lifting 2D image features into 3D space remains a critical challenge. In most previous methods, image features are associated with BEV queries through backward projection~\cite{li2023fb,li2024bevformer,liao2025stcocc}, where a set of 3D reference points is projected onto image planes for feature aggregation. However, these points are predefined and uniformly sampled along globally shared pillars.  
Such a fixed projection space struggles to capture the sparsity and height variations of real-world scenes. As illustrated in Fig.~\ref{Fig:teaser}(a), a car and a tall tree occupy vastly different height ranges. When using a fixed projection space (Fig.~\ref{Fig:teaser}(b)), 3D reference points sampled above the car inevitably misalign with the actual object, leading to the erroneous aggregation of background features (e.g., the sky) instead of the target semantics. Furthermore, even for entirely empty regions devoid of objects, this fixed paradigm still redundantly assigns dense 3D reference points, forcing the network to aggregate meaningless image features. Consequently, this fixed projection space introduces ambiguous correspondences and unreliable feature aggregation.

To address these challenges, we propose HiPR, a camera-LiDAR occupancy framework with Height-Guided Projection Reparameterization. Specifically, as shown in Fig.~\ref{Fig:teaser}(c), we first encode the LiDAR point cloud into a BEV height map, which captures the maximum height of the point cloud. Based on this height prior, we adjust the sampling range of each pillar. As a result, the 3D reference points are redistributed into geometrically meaningful regions rather than fixed height ranges. This enables the projected points to better align with the underlying scene structures in the image space, leading to more consistent and reliable feature aggregation. In addition, for BEV locations without LiDAR heights, we directly mask out the corresponding regions to prevent unreliable feature aggregation. 

Moreover, to alleviate the training instability caused by noisy LiDAR-derived heights, we propose a training-time Progressive Height Conditioning strategy, which gradually transitions the conditioning signal from dense ground-truth heights to LiDAR heights during training.

Our main contributions are summarized as follows:
\begin{itemize}
    \item We introduce a novel camera-LiDAR occupancy paradigm, which, for the first time, leverages the LiDAR height prior to guide the 2D-to-3D view transformation process.
    \item We propose HiPR, which (i) reparameterizes the projection space using a LiDAR-derived height map to align with the scene structure, and (ii) introduces a training-only Progressive Height Conditioning strategy to mitigate the impact of noisy LiDAR heights.
    \item Extensive experiments demonstrate that HiPR achieves state-of-the-art performance. Moreover, a lightweight variant outperforms existing real-time approaches, yielding a favorable trade-off between accuracy and efficiency. Code and weights are released for peer research.
\end{itemize}

\section{Related Work}
\noindent\textbf{Multi-modal 3D Occupancy Prediction.} 
Multi-modal 3D occupancy perception~\cite{duan2025sdgocc,ma2024licrocc,wang2024occgen, wang2023openoccupancy,wei2025ms,wolters2025unleashing,yang2025daocc,zheng2026doracamom} leverages complementary information from multiple sensors to overcome the inherent limitations of single-modality perception. 
A common paradigm is to first lift image features into 3D space and then fuse them with point cloud features. For instance, OccFusion~\cite{ming2024occfusion} concatenates 3D features from different modalities along the feature dimension, followed by multi-scale feature extraction. 
However, the sensor sparsity and modality-specific visibility limit the effectiveness of naive feature concatenation. To address this issue, Co-Occ~\cite{pan2024co} enhances LiDAR features using neighboring camera features, and further introduces an implicit volumetric rendering–based regularization to supervise the fused features. In contrast, DAOcc~\cite{yang2025daocc} introduces 3D object detection as an auxiliary training task, providing additional supervision and improving performance on foreground categories. Beyond fusion strategies based on simple feature concatenation, recent works~\cite{song2025reocc,wei2025ms,zheng2026doracamom} have explored more adaptive fusion mechanisms. For example, REOcc~\cite{song2025reocc} employs cross-attention to fuse camera and radar BEV features. While effective, attention-based fusion introduces considerable computational overhead. To improve efficiency, Occ-Mamba~\cite{qaisar2025occ} uses feature reordering to achieve global modeling with linear complexity. In addition to BEV-level fusion, incorporating 2D perspective-view (PV) features has been shown to further refine 3D representations. For example, HyDRa~\cite{wolters2025unleashing} enhances PV features using BEV point-cloud features through cross-attention mechanisms. 
Unlike these fusion paradigms, we directly exploit LiDAR in the 2D-to-3D view transformation process.

\noindent\textbf{Height Modeling.} 
Height is an important yet often overlooked attribute in BEV perception, providing geometric cues beyond planar layout.
Motivated by the observation that objects are located at distinct height ranges, 
BEV-SAN~\cite{chi2023bev} constructs a height histogram from LiDAR points to determine the boundaries of local BEV slices and then fuses these slices with global BEV features to improve detection accuracy. 
Beyond slice fusion, subsequent methods incorporate height cues directly into the view transformation process. 
OC-BEV~\cite{qi2024ocbev} augments the uniform vertical sampling in BEVFormer~\cite{li2024bevformer} with a scene-level local height prior, thereby focusing sampling on height intervals where objects are most likely to appear. 
Extending this idea to finer granularity, HV-BEV~\cite{wu2025hv} predicts a discrete height distribution for each BEV grid and selects the top-K values for feature sampling. 
Height cues are also important for 3D occupancy prediction, where vertical ambiguities can be more severe. 
For instance, DHD~\cite{wu2025deep} decouples the projection space into multiple subspaces and employs the predicted height map to select features, alleviating height ambiguity. 
DA-Occ~\cite{zhou2025occ} instead leverages the height distribution to aggregate frustum features and further rearranges voxel features to apply direction-aware convolutions. 
Different from these explicit height-supervision paradigms, HBEVOcc~\cite{lyu2026hbevocc} targets the loss of vertical information caused by folding the height dimension into BEV channels and proposes height-aware attention to implicitly capture latent vertical cues within BEV features. In contrast, we explicitly utilize a height map derived from LiDAR to reparameterize the projection space, enabling better alignment with the varying scene structure.

\begin{figure}[t]
    \centering
    \includegraphics[width=1.0\columnwidth]{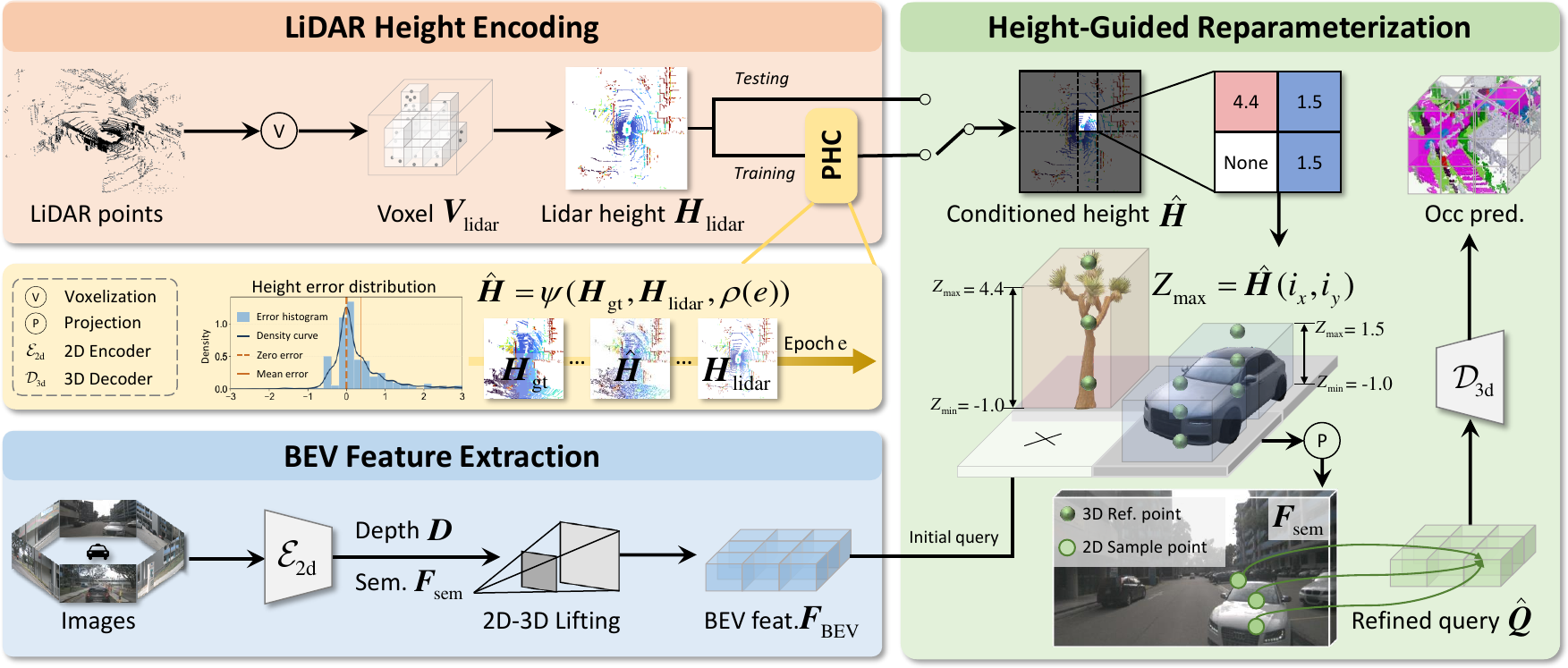}
    \caption{Overview of HiPR. Multi-view images are first transformed into a coarse BEV feature serving as the initial BEV query. Meanwhile, the LiDAR point cloud is encoded into a BEV height map and conditioned by PHC during training. Guided by the height map, the initial BEV query is refined through Height-Guided Reparameterization and finally decoded for 3D occupancy prediction.}
    \vspace{-8pt}
    \label{Fig:pipeline}
\end{figure}

\section{Method}
\subsection{Overview}
As illustrated in Fig.~\ref{Fig:pipeline}, our framework consists of three stages. 
First, multi-view images are processed by the BEV Feature Extraction module, where a 2D encoder generates the depth distribution $\boldsymbol{D}$ and semantic features $\boldsymbol{F}_{\mathrm{sem}}$. These features are lifted into the BEV space through an LSS-based view transformation, yielding a coarse BEV feature $\boldsymbol{F}_{\mathrm{BEV}}$ that serves as the initial BEV query. 
In parallel, the LiDAR point cloud is encoded by the LiDAR Height Encoding module into a BEV height map $\boldsymbol{H}_{\mathrm{lidar}}$, which is further conditioned by Progressive Height Conditioning during training to obtain the conditioned height map $\hat{\boldsymbol{H}}$. 
Next, the Height-Guided Reparameterization module leverages $\hat{\boldsymbol{H}}$ to adaptively reparameterize each BEV pillar and mask out invalid BEV grids, refining the initial BEV query into $\hat{\boldsymbol{Q}}$. Finally, $\hat{\boldsymbol{Q}}$ is fed into a 3D decoder for 3D occupancy prediction.

\subsection{BEV Feature Extraction}
Given multi-view input images, we first extract image features using a 2D encoder (e.g., ResNet~\cite{he2016deep}). For each camera view, the encoder predicts a depth distribution $\boldsymbol{D}$ and a semantic feature map $\boldsymbol{F}_\mathrm{sem}$ from dense image features. The depth distribution represents the probability of each pixel belonging to predefined depth bins, while $\boldsymbol{F}_\mathrm{sem}$ captures appearance and contextual information for scene understanding. Following the LSS-based view transformation~\cite{li2023bevstereo,li2023bevdepth,philion2020lift}, we lift $\boldsymbol{F}_\mathrm{sem}$ from the 2D image plane into the BEV space using $\boldsymbol{D}$ and the camera intrinsic and extrinsic parameters. This process yields a coarse BEV feature, denoted as $\boldsymbol{F}_{\mathrm{BEV}}$, which serves as the initial BEV query for the subsequent Height-Guided Reparameterization module.

\subsection{LiDAR Height Encoding}
\noindent\textbf{Height Map Generation.}
To efficiently encode sparse LiDAR observations, we project the input point cloud into a BEV height map.
Given a point cloud $\mathcal{P}=\{(x_k, y_k, z_k)\}_{k=1}^{K}$ in the ego-vehicle coordinate system, we first define a 3D region of interest bounded within $[x_{\min}, x_{\max})$, $[y_{\min}, y_{\max})$, and $[z_{\min}, z_{\max})$.
The space is then discretized into a voxel grid with horizontal resolution $\Delta_{xy}$ and vertical resolution $\Delta_z$.
Each point $p_k \in \mathcal{P}$ is assigned to a voxel indexed by $(i_x, i_y, i_z)$:
\begin{equation}
\left(i_x, i_y, i_z\right) =
\left(
\left\lfloor\frac{x_k-x_{\min}}{\Delta_{xy}}\right\rfloor,
\left\lfloor\frac{y_k-y_{\min}}{\Delta_{xy}}\right\rfloor,
\left\lfloor\frac{z_k-z_{\min}}{\Delta_z}\right\rfloor
\right).
\label{Eq:voxel_index}
\end{equation}

Based on the voxelized point cloud, we construct a binary occupancy grid $\boldsymbol{V}_{\mathrm{lidar}} \in \{0,1\}^{X \times Y \times Z}$, where $\boldsymbol{V}_{\mathrm{lidar}}(i_x,i_y,i_z)=1$ if the corresponding voxel contains at least one LiDAR point.
To obtain a 2D height representation, we collapse the 3D grid along the vertical axis by selecting the highest occupied voxel index of each non-empty BEV cell $(i_x,i_y)$:
\begin{equation}
i_z^*(i_x, i_y)=\max\left\{i_z \mid \boldsymbol{V}_{\mathrm{lidar}}(i_x,i_y,i_z)=1\right\}.
\label{Eq:max_z}
\end{equation}
We then map the discrete index back to metric height $\boldsymbol{H}_{\mathrm{lidar}} \in \mathbb{R}^{X \times Y}$ as:
\begin{equation}
\boldsymbol{H}_{\mathrm{lidar}}(i_x,i_y)=z_{\min}+\left(i_z^*(i_x,i_y)+1\right)\Delta_z,
\label{Eq:heightmap}
\end{equation}
where the upper boundary of the highest occupied voxel is taken as the representative height. For empty pillars satisfying $\sum_{i_z}\boldsymbol{V}_{\mathrm{lidar}}(i_x,i_y,i_z)=0$, we assign a default invalid value to $\boldsymbol{H}_{\mathrm{lidar}}(i_x,i_y)$.

\noindent\textbf{Progressive Height Conditioning.}
Directly using the LiDAR-derived height map $\boldsymbol{H}_{\mathrm{lidar}}$ for Height-Guided Reparameterization may destabilize training, since $\boldsymbol{H}_{\mathrm{lidar}}$ is inherently noisy, as evidenced by the error distribution in Fig.~\ref{Fig:pipeline}. As the height map directly determines the sampling locations of the subsequent module, such noise can lead to inaccurate sampling.

To mitigate this issue, we introduce a training-only Progressive Height Conditioning (PHC) strategy, which gradually shifts the conditioning signal from a dense ground-truth height map $\boldsymbol{H}_{\mathrm{gt}}$ to the noisy LiDAR height map $\boldsymbol{H}_{\mathrm{lidar}}$. Here, $\boldsymbol{H}_{\mathrm{gt}}$ is derived from the occupancy ground truth using the same height extraction rule as in Eq.~\eqref{Eq:max_z} and \eqref{Eq:heightmap}, ensuring that $\boldsymbol{H}_{\mathrm{gt}}$ and $\boldsymbol{H}_{\mathrm{lidar}}$ are geometrically aligned. During training, the final conditioning map $\hat{\boldsymbol{H}}$ is generated by:
\begin{equation}
\hat{\boldsymbol{H}} = \psi(\boldsymbol{H}_{\mathrm{lidar}}, \boldsymbol{H}_{\mathrm{gt}}, \rho(e)),
\end{equation}
where $\psi(\cdot)$ denotes a stochastic mixing function that independently replaces each valid BEV grid in $\boldsymbol{H}_{\mathrm{lidar}}$ with the corresponding value from $\boldsymbol{H}_{\mathrm{gt}}$ with probability $\rho(e)$, while keeping the remaining grids unchanged. The mixing ratio $\rho(e)$ follows a cosine annealing schedule over $E$ epochs:
\begin{equation}
\rho(e) = \frac{1}{2} \left( 1 + \cos \left( \frac{\pi e}{E} \right) \right).
\label{Eq:cosine_schedule}
\end{equation}
As a result, training starts with strong guidance from $\boldsymbol{H}_{\mathrm{gt}}$ when $\rho(e)=1$, and gradually transitions to the LiDAR-based setting used at inference as $\rho(e)$ decreases to zero. During inference, PHC is disabled and the conditioning map is simply given by $\hat{\boldsymbol{H}}=\boldsymbol{H}_{\mathrm{lidar}}$. In this way, PHC stabilizes early training and improves the robustness of height-guided sampling against noisy LiDAR observations.

\begin{figure}[t]
    \centering
    \includegraphics[width=1.0\columnwidth]{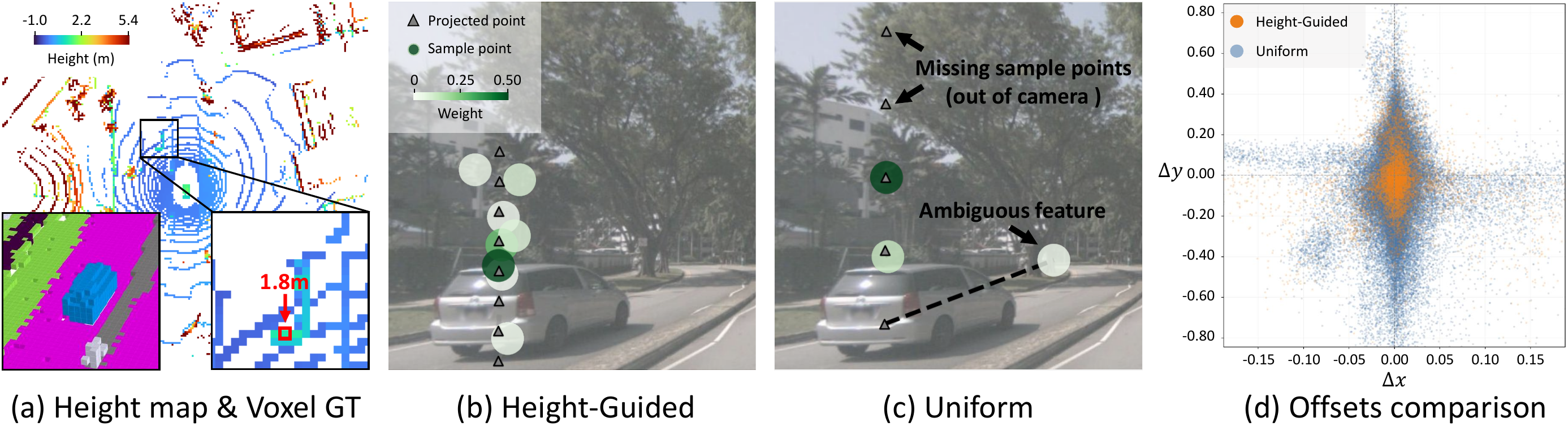}
    \caption{Comparison of height-guided and uniform sampling. (a) Height map and voxel ground truth of a vehicle. (b) Height-guided sampling concentrates projected reference points and sampled features around the valid object region. (c) Uniform sampling introduces invalid projections, including out-of-camera samples and ambiguous feature correspondences. (d) Learned cross-attention offsets show that Height-Guided Reparameterization yields a more compact distribution than uniform sampling.}    
    \label{Fig:HGQR}
    \vspace{-6pt}
\end{figure}

\subsection{Height-Guided Reparameterization}
\noindent\textbf{Revisiting Standard Backward Projection.}
In standard BEVFormer-style backward projection~\cite{li2023fb,li2024bevformer,liao2025stcocc}, each BEV query aggregates image features from a set of 3D reference points sampled uniformly along a globally fixed pillar:
\begin{equation}
\hat{\boldsymbol{Q}}(i_x, i_y)=\sum_{j=1}^{N_z} \phi\left(\boldsymbol{Q}(i_x, i_y), \mathcal{R}\left(i_x, i_y, z_j\right), \boldsymbol{F}_\mathrm{sem}\right),
\end{equation}
where $\boldsymbol{Q}(i_x, i_y)$ and $\hat{\boldsymbol{Q}}(i_x, i_y)$ denote the input and updated BEV queries at location $(i_x, i_y)$, $\phi(\cdot)$ represents the deformable cross-attention operator~\cite{li2024bevformer}, $\mathcal{R}\left(i_x, i_y, z_j\right)$ stands for the 3D reference point at height $z_j$, $\boldsymbol{F}_\mathrm{sem}$ refers to the multi-view image features, and $N_z$ is the number of vertical sampling points. The vertical sampling locations~\cite{li2023fb,li2024bevformer,liao2025stcocc} are uniformly defined as:
\begin{equation}
z_j = z_{\min}+\alpha_j(z_{\max}-z_{\min}), \qquad \alpha_j=\frac{j-1}{N_z-1}.
\end{equation}
Since the vertical sampling locations $\{z_j\}_{j=1}^{N_z}$ are identical for all BEV grids, this formulation is inherently agnostic to scene geometry. As a result, reference points are often wasted in empty spaces, and even for object regions, the uniform sampling causes their 2D projections to misalign with the actual targets, leading to ambiguous correspondences and unreliable feature aggregation.

\noindent\textbf{Reparameterization.}
To deal with this problem, we propose utilizing the conditioned height map $\hat{\boldsymbol{H}}(i_x,i_y)$ to dynamically reparameterize the projection space of each BEV query. 

Specifically, we define a height-validity mask:
\begin{equation}
\boldsymbol{M}(i_x,i_y)=\mathbb{I}\big(\hat{\boldsymbol{H}}(i_x,i_y)\ \text{is valid}\big),
\end{equation}
and construct height-guided sampling locations as:
\begin{equation}
\tilde z_j(i_x,i_y)=z_{\min}+\alpha_j\big(\hat{\boldsymbol{H}}(i_x,i_y)-z_{\min}\big), \qquad \alpha_j=\frac{j-1}{N_z-1}.
\end{equation}
Accordingly, the query update process is reformulated as:
\begin{equation}
\hat{\boldsymbol{Q}}(i_x, i_y)=
\begin{cases}
\displaystyle \sum_{j=1}^{N_z}\phi\!\left(\boldsymbol{Q}(i_x, i_y), \mathcal{R}\!\left(i_x, i_y, \tilde z_j(i_x,i_y)\right), \boldsymbol{F}_\mathrm{sem}\right), & \boldsymbol{M}(i_x,i_y)=1,\\[6pt]
\boldsymbol{Q}(i_x,i_y), & \boldsymbol{M}(i_x,i_y)=0.
\end{cases}
\end{equation} 
In this way, the conditioned height map $\hat{\boldsymbol{H}}$ defines the upper height bound of each pillar and restricts feature aggregation to valid BEV regions. The effectiveness of this reparameterization is illustrated in Fig.~\ref{Fig:HGQR}. Consider a target vehicle with a physical height of $1.8$~m, as indicated by the LiDAR height map in Fig.~\ref{Fig:HGQR}(a). Under the standard uniform sampling paradigm (Fig.~\ref{Fig:HGQR}(c)), the model samples reference points up to the maximum predefined ceiling (e.g., $z_{\max}=5.4$~m). This lack of geometric awareness results in numerous invalid projections that either fall out of the camera field of view or sample ambiguous background features (e.g., trees). In contrast, by dynamically truncating the sampling range to the actual object height ($[z_{\min}, 1.8]$~m), our height-guided method encourages the sampling points to concentrate on the semantically meaningful vehicle body (Fig.~\ref{Fig:HGQR}(b)). 
Furthermore, as shown in Fig.~\ref{Fig:HGQR}(d), the learned spatial offsets $(\Delta x, \Delta y)$ of our method exhibit a compact, zero-centered distribution. This suggests that the initial 3D projections are well aligned with informative regions and require only minor adjustments. In contrast, uniform sampling produces a more scattered distribution, indicating that the model relies on larger offsets to capture valid features.

\subsection{Loss Function}
Following~\cite{chen2025alocc,cheng2022masked,liu2024fully}, we train the query-based decoder with a classification loss, a mask loss, and a dice loss. The overall objective is defined as:
\begin{equation}
\mathcal{L} = \lambda_{\mathrm{cls}} \mathcal{L}_{\mathrm{cls}} +
\lambda_{\mathrm{mask}} \mathcal{L}_{\mathrm{mask}} +
\lambda_{\mathrm{dice}} \mathcal{L}_{\mathrm{dice}},
\end{equation}
where $\mathcal{L}_{\mathrm{cls}}$ denotes the cross-entropy loss for query classification, $\mathcal{L}_{\mathrm{mask}}$ denotes the cross-entropy loss for occupancy masks, and $\mathcal{L}_{\mathrm{dice}}$ denotes the dice loss for mask overlap optimization. In our implementation, the loss weights are set to $\lambda_{\mathrm{cls}} = 2.0$, $\lambda_{\mathrm{mask}} = 5.0$, and $\lambda_{\mathrm{dice}} = 5.0$.


\begin{table*}[!t]
\centering
\Huge
\caption{Performance on Occ3D. The camera visible mask is used during training. Inputs include Camera (C), Radar (R) and LIDAR (L). R denotes ResNet, and Swin denotes Swin Transformer.}
\vspace{-6pt}
\renewcommand\arraystretch{1.2}
\resizebox{1.0\textwidth}{!}{
\begin{tabular}{l|cc|c|ccccccccccccccccc}

\toprule
Method
& Input
& Backbone
& mIoU
& \begin{sideways}{\textcolor{otherscolor}{$\blacksquare$} others}\end{sideways}  
& \begin{sideways}{\textcolor{barriercolor}{$\blacksquare$} barrier}\end{sideways}  
& \begin{sideways}{\textcolor{bicyclecolor}{$\blacksquare$} bicycle}\end{sideways}  
& \begin{sideways}{\textcolor{buscolor}{$\blacksquare$} bus}\end{sideways}  
& \begin{sideways}{\textcolor{carcolor}{$\blacksquare$} car}\end{sideways}  
& \begin{sideways}{\textcolor{constructcolor}{$\blacksquare$} cons. veh.}\end{sideways}  
& \begin{sideways}{\textcolor{motorcolor}{$\blacksquare$} motorcycle}\end{sideways}  
& \begin{sideways}{\textcolor{pedestriancolor}{$\blacksquare$} pedestrian}\end{sideways}  
& \begin{sideways}{\textcolor{trafficcolor}{$\blacksquare$} traffic cone}\end{sideways}  
& \begin{sideways}{\textcolor{trailercolor}{$\blacksquare$} trailer}\end{sideways}  
& \begin{sideways}{\textcolor{truckcolor}{$\blacksquare$} truck}\end{sideways}  
& \begin{sideways}{\textcolor{drivablecolor}{$\blacksquare$} drive. surf.}\end{sideways}  
& \begin{sideways}{\textcolor{otherflatcolor}{$\blacksquare$} other flat}\end{sideways}  
& \begin{sideways}{\textcolor{sidewalkcolor}{$\blacksquare$} sidewalk}\end{sideways}  
& \begin{sideways}{\textcolor{terraincolor}{$\blacksquare$} terrain}\end{sideways}  
& \begin{sideways}{\textcolor{manmadecolor}{$\blacksquare$} manmade}\end{sideways}  
& \begin{sideways}{\textcolor{vegetationcolor}{$\blacksquare$} vegetation}\end{sideways} 
\\ 
\midrule

BEVDetOcc~\cite{huang2021bevdet} & C & Swin-B & 42.0  & 12.2  & 49.6  & 25.1  & 52.0  & 54.5  & 27.9  & 28.0  & 28.9  & 27.2  & 36.4  & 42.2  & 82.3  & 43.3  & 54.6  & 57.9  & 48.6  & 43.6 \\ 
FB-Occ~\cite{li2023fb} & C & R50 & 39.8  & 13.8  & 44.5  & 27.1  & 46.2  & 49.7  & 24.6  & 27.4  & 28.5  & 28.2  & 33.7  & 36.5  & 81.7  & 44.1  & 52.6  & 56.9  & 42.6  & 38.1  \\ 
FlashOcc~\cite{yu2023flashocc} & C & R50 & 32.0  & 6.2  & 39.6  & 11.3  & 36.3  & 44.0  & 16.3  & 14.7  & 16.9  & 15.8  & 28.6  & 30.9  & 78.2  & 37.5  & 47.4  & 51.4  & 36.8  & 31.4  \\ 
COTR~\cite{ma2024cotr} & C & R50 & 44.5  & 13.3  & 52.1  & 32.0  & 46.0  & 55.6  & 32.6  & 32.8  & 30.4  & 34.1  & 37.7  & 41.8  & 84.5  & 46.2  & 57.6  & 60.7  & 52.0  & 46.3 \\
ProtoOcc~\cite{kim2025protoocc} & C & R50 & 39.6  & 12.4  & 45.9  & 26.3  & 44.4  & 51.8  & 26.8  & 27.6  & 28.0  & 27.5  & 32.8  & 36.9  & 81.8  & 45.7  & 53.1  & 56.5  & 42.2  & 36.6  \\ 
STCOcc~\cite{liao2025stcocc} & C & R50 & 45.0  & 15.2  & 52.3  & 32.2  & 50.5  & 56.5  & 31.7  & 33.9  & 33.4  & 33.8  & 38.9  & 44.9  & 83.9  & 47.4  & 57.1  & 60.1  & 50.6  & 42.7 \\ 
RadOcc~\cite{zhang2024radocc} & C+L & Swin-B & 49.4  & 10.9  & 58.2  & 25.0  & 57.9  & 62.9  & 34.0  & 33.5  & 50.1  & 32.1  & 48.9  & 52.1  & 82.9  & 42.7  & 55.3  & 58.3  & 68.6  & 66.0   \\ 
OccFusion~\cite{ming2024occfusion} & C+L+R & R101 & 46.7  & 12.4  & 50.3  & 31.5  & 57.6  & 58.8  & 34.0  & 41.0  & 47.2  & 29.7  & 42.0  & 48.0  & 78.4  & 35.7  & 47.3  & 52.7  & 63.5  & 63.3  \\
HyDRa~\cite{wolters2025unleashing} & C+R & R50 & 44.4  & 15.1  & 51.1  & 32.7  & 52.3  & 56.3  & 29.4  & 35.9  & 35.1  & 33.7  & 39.1  & 44.1  & 80.4  & 45.1  & 52.0  & 55.3  & 52.1  & 44.4  \\ 
EFFOcc~\cite{shi2025effocc} & C+L & R50 & 52.8  & 12.1  & 59.7  & 33.4  & \second{61.8}  & \second{65.0}  & 35.5  & 46.0  & 57.1  & 41.0  & 47.9  & 54.6  & 82.8  & 44.0  & 56.4  & 60.2  & \first{71.1}  & \first{69.6}  \\ 
SDG-Fusion~\cite{duan2025sdgocc} & C+L & R50 & 51.7  & 13.2  & 57.8  & 24.3  & 60.3  & 64.3  & 36.2  & 39.4  & 52.4  & 35.8  & \first{50.9}  & 53.7  & 84.6  & 47.5  & 58.0  & 61.6  & \second{70.7}  & 67.7  \\
SDG-KL~\cite{duan2025sdgocc} & C+L & R50 & 50.2  & 12.3  & 57.1  & 23.7  & 58.8  & 62.7  & 34.6  & 36.2  & 50.1  & 32.1  & 49.9  & 51.2  & 84.1  & 46.1  & 57.2  & 61.5  & 69.6  & 65.8 \\ 
ALOcc-2D-mini~\cite{chen2025alocc} & C+L & R50 & 50.0  & 15.7  & 54.6  & 36.6  & 55.7  & 60.6  & 34.8  & 41.0  & 44.9  & 39.3  & 44.5  & 51.1  & 83.6  & 48.5  & 57.3  & 60.2  & 62.7  & 58.2  \\
ALOcc-2D~\cite{chen2025alocc}  & C+L & R50 & 53.5  & \second{16.5}  & 57.8  & \first{41.6}  & 57.9  & 63.8  & 37.6  & 45.0  & 52.1  & \second{45.8}  & 49.6  & 54.4  & \first{85.3}  & \second{50.5}  & \second{59.7}  & \second{62.3}  & 67.1  & 62.0 \\
DAOcc~\cite{yang2025daocc} & C+L & R50 & \second{54.3}  & 13.0  & \first{60.7}  & 39.8  & \first{64.0}  & \first{66.5}  & 36.3  & \first{49.0}  & \first{60.1}  & 44.3  & \second{50.7}  & \first{55.9}  & 82.9  & 44.6  & 56.8  & 60.6  & 70.1  & \second{68.3}  \\
\hline
\textbf{HiPR-mini} & C+L & R50 & 53.1  & \first{17.0}  & 58.8  & 38.8  & 58.0  & 62.8  & \second{37.7}  & 45.4  & 55.7  & 43.1  & 49.3  & 53.5  & 83.9  & 49.3  & 57.9  & 60.7  & 66.9  & 63.8  \\ 
\textbf{HiPR} & C+L & R50 & \first{54.7}  & 16.4  & \second{59.7}  & \second{40.4}  & 59.7  & 64.9  & \first{38.9}  & \second{47.8}  & \second{57.5}  & \first{46.2}  & 48.9  & \second{55.3}  & \second{85.2}  & \first{51.6}  & \first{59.9}  & \first{62.7}  & 69.1  & 65.4   \\ 

\bottomrule
\end{tabular}}
\label{tab:Occ3d_w_camera_mask}
\vspace{-3pt}
\end{table*}

\section{Experiments}
\label{Sec:Exp}
\noindent\textbf{Datasets and Metrics.} 
We evaluate our method on the nuScenes dataset~\cite{caesar2020nuscenes}, which contains 700 training scenes and 150 validation scenes with annotations at 2~Hz. Based upon nuScenes, Occ3D~\cite{tian2023occ3d} and SurroundOcc~\cite{wei2023surroundocc} further provide voxel-wise occupancy annotations. For Occ3D, the voxel region spans $[-40, 40]$~m along the $X$ and $Y$ axes and $[-1, 5.4]$~m along the $Z$ axis, with a voxel size of $0.4 \times 0.4 \times 0.4$~m. It contains 18 semantic categories, including 17 object classes and 1 empty class. For SurroundOcc, the voxel region is $[-50, 50]$~m along the $X$ and $Y$ axes and $[-5, 3]$~m along the $Z$ axis, with a voxel size of $0.5 \times 0.5 \times 0.5$~m. Compared with Occ3D, SurroundOcc excludes the ``others'' category. Following prior work~\cite{liao2025stcocc,liu2024fully,yang2025daocc}, we use mIoU and RayIoU as evaluation metrics.

\noindent\textbf{Implementation Details.}
Built upon the ALOcc series~\cite{chen2025alocc}, we develop two variants: a lightweight \textbf{HiPR-mini} for real-time performance, and a standard \textbf{HiPR} that incorporates stereo depth estimation and larger channel dimensions. We optimize all models using AdamW~\cite{loshchilov2017decoupled} with an initial learning rate of $2\times10^{-4}$ for 24 epochs and a total batch size of 16. Our implementation is built on MMDetection3D~\cite{chen2019mmdetection}. Experiments are conducted on NVIDIA GeForce RTX 4090 and L40X GPUs.

\begin{figure}[!t]
    \centering

    \begin{minipage}{0.52\textwidth}
        \centering
        \captionof{table}{Comparison of RayIoU on Occ3D. The camera visible mask is not used during the training.}
        \vspace{-6pt}
        \renewcommand\arraystretch{1.15}
        \resizebox{\textwidth}{!}{
            \begin{tabular}{l|cc|c|ccc}
                \toprule
                Method & Input & Backbone & RayIoU & \multicolumn{3}{c}{RayIoU\textsubscript{1m, 2m, 4m}} \\ 
                \midrule
                BEVFormer~\cite{li2024bevformer} & C  & R101 & 32.4  & 26.1  & 32.9  & 38.0  \\ 
                FB-Occ~\cite{li2023fb} & C &  R50 & 39.0  & 33.0  & 40.0  & 44.0  \\ 
                RenderOcc~\cite{pan2024renderocc} & C  & Swin-B & 24.4  & 13.4  & 19.6  & 25.5   \\ 
                SparseOcc~\cite{liu2024fully} & C  & R50 & 36.1  & 30.2  & 36.8  & 41.2    \\ 
                FlashOcc~\cite{yu2023flashocc} & C & R50 & 38.5  & 32.8  & 39.3  & 43.4    \\ 
                OPUS~\cite{wang2024opus} & C  & R50 & 41.2  & 34.7  & 42.1  & 46.7   \\
                STCOcc~\cite{liao2025stcocc} & C & R50 & 42.1  & 36.9  & 42.8  & 46.7  \\ 
                ALOcc-2D-mini~\cite{chen2025alocc} & C & R50 & 39.3  & 32.9  & 40.1  & 44.8   \\ 
                ALOcc-2D~\cite{chen2025alocc} & C & R50 & 43.0  & 37.1  & 43.8  & 48.2   \\ 
                DAOcc~\cite{yang2025daocc} & C+L & R50 & \second{48.4}  & \second{44.5}  & 48.9  & 51.9   \\ 
                \hline
                \textbf{HiPR-mini} & C+L  & R50 & \second{48.4}  & 44.2  & \second{49.0}  & \second{52.0}  \\ 
                \textbf{HiPR} & C+L  & R50 & \first{50.0}  & \first{45.9}  & \first{50.5} & \first{53.4}    \\ 
                \bottomrule
            \end{tabular}
            \label{tab:occ3d_womask_rayiou}
        }
    \end{minipage}
    \hfill
    \begin{minipage}{0.45\textwidth}
        \centering
        \includegraphics[width=\textwidth]{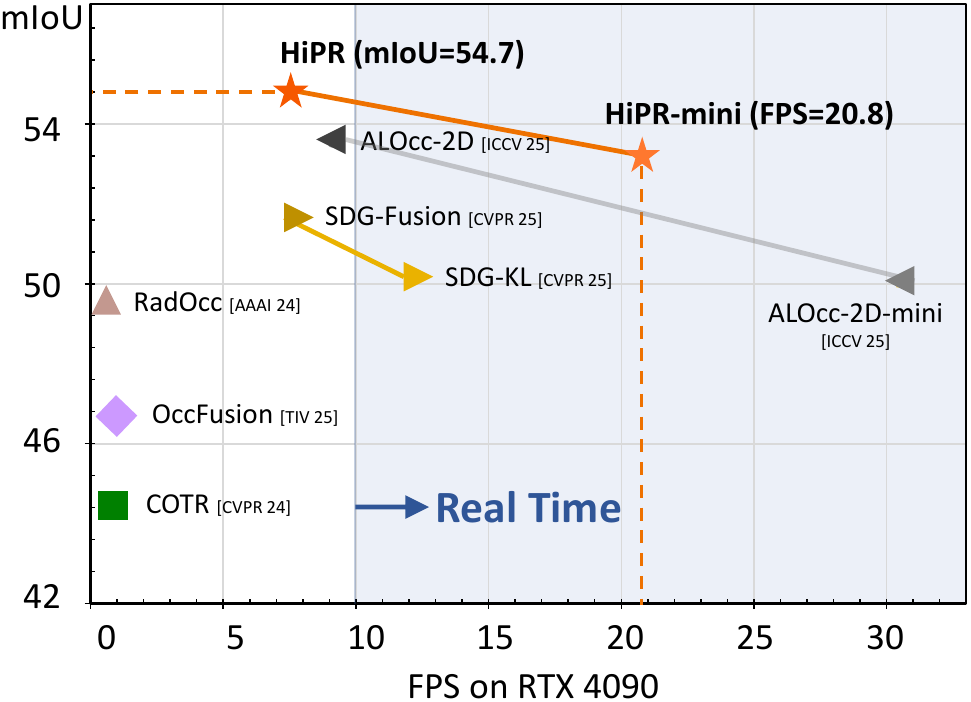}
        \vspace{-15pt}
        \caption{FPS-mIoU comparison.}
        \label{Fig:fps}
    \end{minipage}
\vspace{-5pt}
\end{figure}

\begin{table*}[!t]
\centering
\Huge
\caption{Performance on SurroundOcc. Unlike Occ3D, the evaluation is performed over all voxels.}
\vspace{-6pt}
\renewcommand\arraystretch{1.15}
\resizebox{1.0\textwidth}{!}{
\begin{tabular}{l|cc|c|cccccccccccccccc}
\toprule
Method
& Input
& Backbone
& mIoU
& \begin{sideways}{\textcolor{barriercolor}{$\blacksquare$} barrier}\end{sideways}  
& \begin{sideways}{\textcolor{bicyclecolor}{$\blacksquare$} bicycle}\end{sideways}  
& \begin{sideways}{\textcolor{buscolor}{$\blacksquare$} bus}\end{sideways}  
& \begin{sideways}{\textcolor{carcolor}{$\blacksquare$} car}\end{sideways}  
& \begin{sideways}{\textcolor{constructcolor}{$\blacksquare$} cons. veh.}\end{sideways}  
& \begin{sideways}{\textcolor{motorcolor}{$\blacksquare$} motorcycle}\end{sideways}  
& \begin{sideways}{\textcolor{pedestriancolor}{$\blacksquare$} pedestrian}\end{sideways}  
& \begin{sideways}{\textcolor{trafficcolor}{$\blacksquare$} traffic cone}\end{sideways}  
& \begin{sideways}{\textcolor{trailercolor}{$\blacksquare$} trailer}\end{sideways}  
& \begin{sideways}{\textcolor{truckcolor}{$\blacksquare$} truck}\end{sideways}  
& \begin{sideways}{\textcolor{drivablecolor}{$\blacksquare$} drive. surf.}\end{sideways}  
& \begin{sideways}{\textcolor{otherflatcolor}{$\blacksquare$} other flat}\end{sideways}  
& \begin{sideways}{\textcolor{sidewalkcolor}{$\blacksquare$} sidewalk}\end{sideways}  
& \begin{sideways}{\textcolor{terraincolor}{$\blacksquare$} terrain}\end{sideways}  
& \begin{sideways}{\textcolor{manmadecolor}{$\blacksquare$} manmade}\end{sideways}  
& \begin{sideways}{\textcolor{vegetationcolor}{$\blacksquare$} vegetation}\end{sideways} \\
\midrule

BEVFormer~\cite{li2024bevformer} & C & R101 & 16.8  & 14.2  & 6.6  & 23.5  & 28.3  & 8.7  & 10.8  & 6.6  & 4.1  & 11.2  & 17.8  & 37.3  & 18.0  & 22.9  & 22.2  & 13.8  & 22.2  \\
TPVFormer~\cite{huang2023tri} & C & R101 & 17.1  & 16.0  & 5.3  & 23.9  & 27.3  & 9.8  & 8.7  & 7.1  & 5.2  & 11.0  & 19.2  & 38.9  & 21.3  & 24.3  & 23.2  & 11.7  & 20.8   \\ 
SurroundOcc~\cite{wei2023surroundocc} & C & R101 & 20.3  & 20.6  & 11.7  & 28.1  & 30.9  & 10.7  & 15.1  & 14.1  & 12.1  & 14.4  & 22.3  & 37.3  & 23.7  & 24.5  & 22.8  & 14.9  & 21.9   \\ 
GaussianFormer~\cite{huang2024gaussianformer} & C & R101 & 17.3  & 19.5  & 11.3  & 26.1  & 29.8  & 10.5  & 13.8  & 12.6  & 8.7  & 12.7  & 21.6  & 39.6  & 23.3  & 24.5  & 23.0  & 9.6  & 19.1   \\ 
GaussianFormer2~\cite{huang2025gaussianformer}& C & R101 & 20.8  & 21.4  & 13.4  & 28.5  & 30.8  & 10.9  & 15.8  & 13.6  & 10.5  & 14.0  & 22.9  & 40.6  & 24.4  & 26.1  & 24.3  & 13.8  & 22.0   \\ 

M-CONet~\cite{wang2023openoccupancy} & C+L & R101 & 24.7  & 24.8  & 13.0  & 31.6  & 34.8  & 14.6  & 18.0  & 20.0  & 14.7  & 20.0  & 26.6  & 39.2  & 22.8  & 26.1  & 26.0  & 26.0  & 37.1   \\
Co-Occ~\cite{pan2024co} & C+L & R101 & 27.1  & 28.1  & 16.1  & 34.0  & \second{37.2}  & 17.0  & 21.6  & 20.8  & 15.9  & 21.9  & 28.7  & 42.3  & 25.4  & \second{29.1}  & \second{28.6}  & 28.2  & 38.0   \\
OccFusion~\cite{ming2024occfusion} & C+L & R101 & 27.6  & 25.2  & 19.9  & 34.8  & 36.2  & 20.0  & 23.1  & 25.3  & 17.5  & \first{22.7}  & 30.1  & 39.5  & 23.3  & 25.7  & 27.6  & 29.5  & 40.6   \\ 
OccFusion~\cite{ming2024occfusion} & C+L+R & R101 & 28.3  & \second{28.3}  & 21.0  & \first{35.1}  & 36.8  & 20.3  & 26.2  & 25.9  & 19.2  & 21.3  & \second{30.6}  & 40.1  & 23.8  & 25.6  & 27.6  & 29.8  & \second{40.8}   \\ 
OccCylindrical~\cite{ming2025occcylindrical} & C+L & R50 & \second{28.7}  & 26.2  & \second{22.1}  & 31.5  & 36.8  & 18.0  & \second{27.8}  & \first{29.9}  & \first{23.9}  & 20.6  & 28.3  & \second{43.0}  & 23.1  & 28.0  & 27.8  & \first{30.8}  & \first{41.0}   \\ 
\hline
\textbf{HiPR-mini} & C+L & R50 & 28.6  & 27.9  & 21.7  & 33.2  & 36.8  & \second{22.8}  & 27.7  & 27.0  & 19.9  & 20.2  & 30.0  & 41.4  & \second{26.8}  & 28.7  & 27.7  & 28.3  & 37.5   \\ 
\textbf{HiPR} & C+L & R50 & \first{30.4}  & \first{29.9}  & \first{23.1}  & \second{34.8}  & \first{38.4}  & \first{23.4}  & \first{29.7}  & \second{28.8}  & \second{21.9}  & \second{22.4}  & \first{32.0}  & \first{43.8}  & \first{28.6}  & \first{30.6}  & \first{29.9}  & \second{30.6}  & 39.4   \\ 

\bottomrule
\end{tabular}}
\label{tab:surroundocc_res}

\end{table*}

\subsection{Comparison with the State-of-the-Art}

\begin{figure}[!t]
    \centering
    \includegraphics[width=1.0\columnwidth]{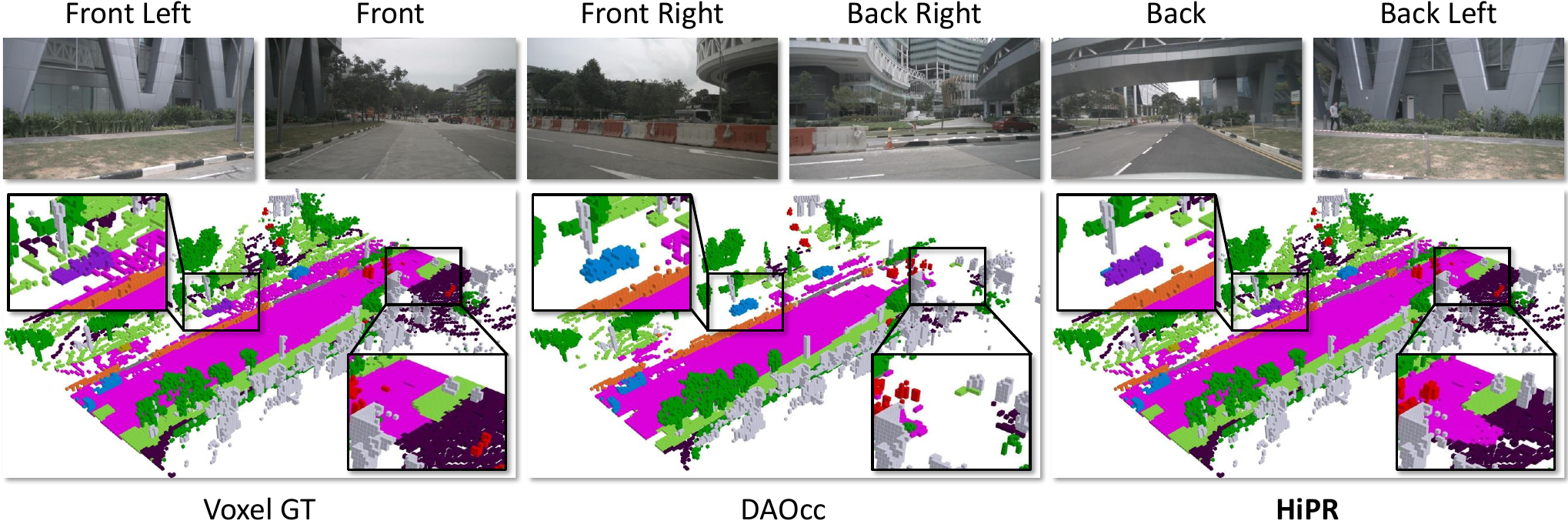}
    \caption{Visual comparisons of DAOcc and our HiPR on the validation set of Occ3D. }
    \vspace{0pt}
    \label{Fig:vis_occ_compare_womask}
\end{figure}

\noindent\textbf{Comparison on Occ3D.} 
As shown in Tab.~\ref{tab:Occ3d_w_camera_mask} and Tab.~\ref{tab:occ3d_womask_rayiou}, we evaluate HiPR on Occ3D under both training settings: with and without the camera mask. Overall, our method achieves state-of-the-art performance in both scenarios. Under the \textit{with camera mask} setting, we report mIoU. HiPR outperforms the second-best DAOcc~\cite{yang2025daocc} by 0.4 mIoU. Compared to OccFusion~\cite{ming2024occfusion}, which leverages additional radar input and a larger backbone (ResNet-101~\cite{he2016deep}), HiPR achieves a substantial gain of 8.0 mIoU. Under the \textit{without camera mask} setting, we report RayIoU~\cite{liu2024fully}. HiPR significantly outperforms camera-based methods such as STCOcc~\cite{liao2025stcocc} by a large margin of 7.9 RayIoU. Compared to the multi-modal method DAOcc, the lightweight variant HiPR-mini achieves comparable performance, while the full HiPR further improves upon it by 1.6 RayIoU. 
Fig.~\ref{Fig:vis_occ_compare_womask} shows qualitative comparisons between HiPR and DAOcc. DAOcc misclassifies large objects and fails to reconstruct distant structures accurately, while HiPR yields more accurate semantics and more complete reconstructions.

\noindent\textbf{Comparison on SurroundOcc.}
Tab.~\ref{tab:surroundocc_res} presents a quantitative comparison on the SurroundOcc~\cite{wei2023surroundocc} dataset. Overall, HiPR consistently outperforms existing methods. In particular, it surpasses the second-best method, OccCylindrical~\cite{ming2025occcylindrical}, by 1.7 mIoU. Meanwhile, despite using only a lightweight ResNet-50~\cite{he2016deep} backbone, HiPR still outperforms competitive approaches such as OccFusion~\cite{ming2024occfusion}, Co-Occ~\cite{pan2024co}, and GaussianFormer~\cite{huang2024gaussianformer}. Furthermore, HiPR achieves clear improvements on challenging categories such as bicycle and motorcycle, which are typically sparse and difficult to model, thereby demonstrating its ability to better capture fine-grained structures.

\noindent\textbf{Speed--Accuracy Trade-off.}
Fig.~\ref{Fig:fps} presents the speed--accuracy trade-off of HiPR on Occ3D under the setting trained without camera visible mask. Following prior works~\cite{he2025achieving,hou2024fastocc,kim2025protoocc}, methods running above 10 FPS are regarded as real-time. 
Overall, HiPR attains state-of-the-art accuracy, while HiPR-mini enables real-time inference with near state-of-the-art performance, outperforming existing feature-fusion-based multi-modal methods~\cite{duan2025sdgocc, ming2024occfusion,zhang2024radocc}. Specifically, compared with the real-time method SDG-KL~\cite{duan2025sdgocc}, HiPR-mini is 7.2 FPS faster and improves mIoU by 2.9 points, demonstrating its superior speed--accuracy trade-off for efficient 3D occupancy prediction.

\begin{figure*}[!t]
\centering

\begin{minipage}[t]{0.59\textwidth}
    \vspace{0pt}
    \centering
    \captionof{table}{Ablation study of HiPR on the Occ3D dataset. HGR refers to the Height-Guided Reparameterization and PHC refers to the Progressive Height Conditioning.}
    \vspace{-2pt}
    \label{tab:ablation}
    \resizebox{\textwidth}{!}{
    \begin{tabular}{lcccc}
    \toprule
    \multirow{2}{*}{Model} & \multicolumn{2}{c}{HGR} & \multirow{2}{*}{PHC} & \multirow{2}{*}{mIoU} \\
    \cmidrule(lr){2-3}
    & \makecell{Height-Guided Sampling} & \makecell{Height-Validity Mask} &  &  \\
    \midrule
    Baseline &  &  &  & 50.01 \\
    i   & $\checkmark$ &  &  & 51.04 \\
    ii  & $\checkmark$ & $\checkmark$ &  & 52.81 \\
    iii & $\checkmark$ & $\checkmark$ & $\checkmark$ & 53.10 \\
    \bottomrule
    \end{tabular}
    }
\end{minipage}
\hfill
\begin{minipage}[t]{0.37\textwidth}
    \vspace{0pt}
    \centering
    \includegraphics[width=\textwidth]{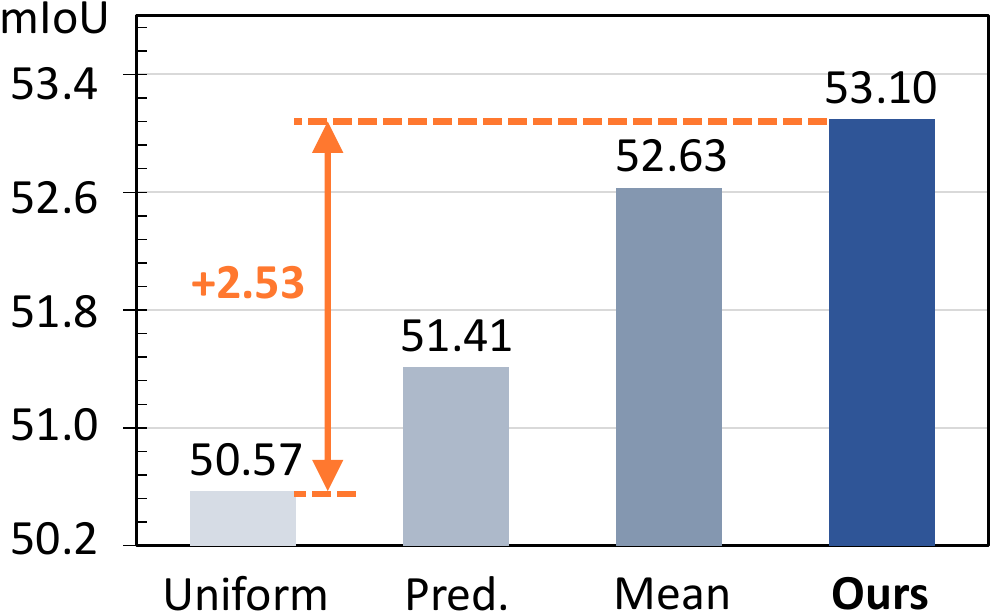}
    \vspace{-15pt}
    \captionof{figure}{Ablation on HGR.}
    \label{Fig:ablation_HGR}
    \vspace{-8pt}
\end{minipage}

\end{figure*}

\subsection{Ablation Study}
All experiments in this section are trained with the camera visible mask on the Occ3D dataset.

\begin{figure}[!t]
    \centering
    \begin{minipage}{0.62\columnwidth}
        \vspace{0pt}
        \centering
        \includegraphics[width=\linewidth]{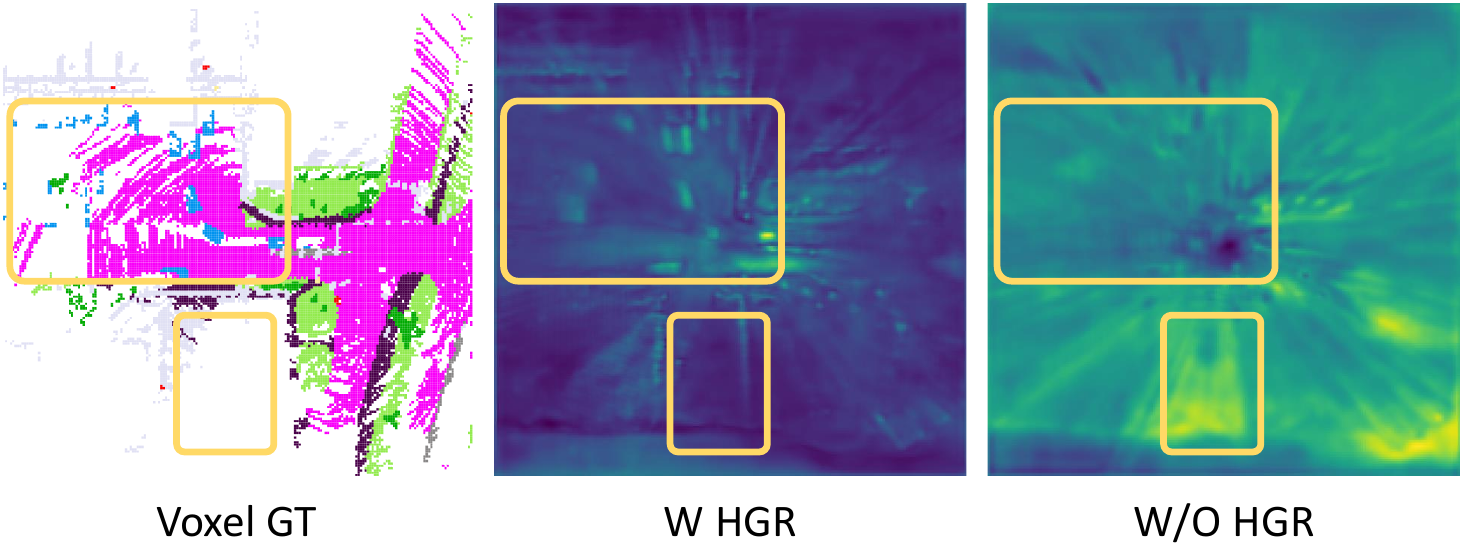}
        \caption{Comparison of BEV feature with and without HGR.}
        \label{Fig:bev_feat_compare}
    \end{minipage}
    \hfill
    \begin{minipage}{0.33\columnwidth}
        \vspace{0pt}
        \centering
        \captionof{table}{Ablation study of conditioning strategy in PHC module.}
        \label{tab:cond}
        \begin{tabular}{lc}
        \toprule
        Setting & mIoU \\
        \midrule
        LiDAR Only & 52.81 \\
        GT Only & 31.83 \\
        Linear Interpolation & 52.83 \\
        Replace (\textbf{Ours}) & 53.10 \\
        \bottomrule
        \end{tabular}
    \end{minipage}
\end{figure}

\noindent\textbf{HiPR Designs.} 
Tab.~\ref{tab:ablation} presents the ablation study of the proposed Height-Guided Reparameterization (HGR) and Progressive Height Conditioning (PHC). The baseline model is ALOcc-2D-mini~\cite{chen2025alocc}. Starting from this baseline, introducing height-guided sampling improves the mIoU from 50.01 to 51.04. In this setting, valid pillars are reparameterized according to the height map, while invalid regions still use uniformly sampled pillars over the maximum height range. Further incorporating the height-validity mask brings a larger improvement, boosting the mIoU to 52.81. This demonstrates that restricting feature aggregation to valid BEV regions is crucial for suppressing invalid projections and improving geometric alignment. As shown in Fig.~\ref{Fig:bev_feat_compare}, the resulting BEV features are more focused and semantically consistent in occupied regions, while noisy responses in empty areas are effectively suppressed. Finally, PHC further improves the performance by 0.29 mIoU, indicating that progressively transitioning stabilizes training and improves the robustness of height-guided sampling.

\noindent\textbf{Sampling Strategies in HGR.}
Fig.~\ref{Fig:ablation_HGR} compares different height sampling strategies in HGR. Starting from HiPR-mini, we keep all other components fixed and only replace the height-guided sampling strategy. Using a uniform sampling range from $z_{\min}$ to $z_{\max}$ achieves 50.57 mIoU. Replacing it with a learned height predictor improves performance to 51.41 mIoU, demonstrating the benefit of adaptive height modeling. Specifically, in this setting, we remove the LiDAR branch and append a height prediction head to the BEV features, supervised by a cross-entropy loss. The improvement over uniform sampling indicates that fixed sampling is suboptimal for scenes with varying local geometry, while adaptive height estimation provides a more suitable sampling prior. Introducing LiDAR geometry further improves performance. Using the mean pillar height reaches 52.63 mIoU, while our strategy of using the highest LiDAR-observed height achieves the best result of 53.10 mIoU. These results show that LiDAR-guided height sampling is more effective than learned or uniform alternatives, and that the highest occupied height provides the most informative prior for HGR.

\begin{figure}[!t]
\centering
\begin{minipage}[t]{0.378\linewidth}
    \vspace{0pt}
    \centering
    \includegraphics[width=\linewidth]{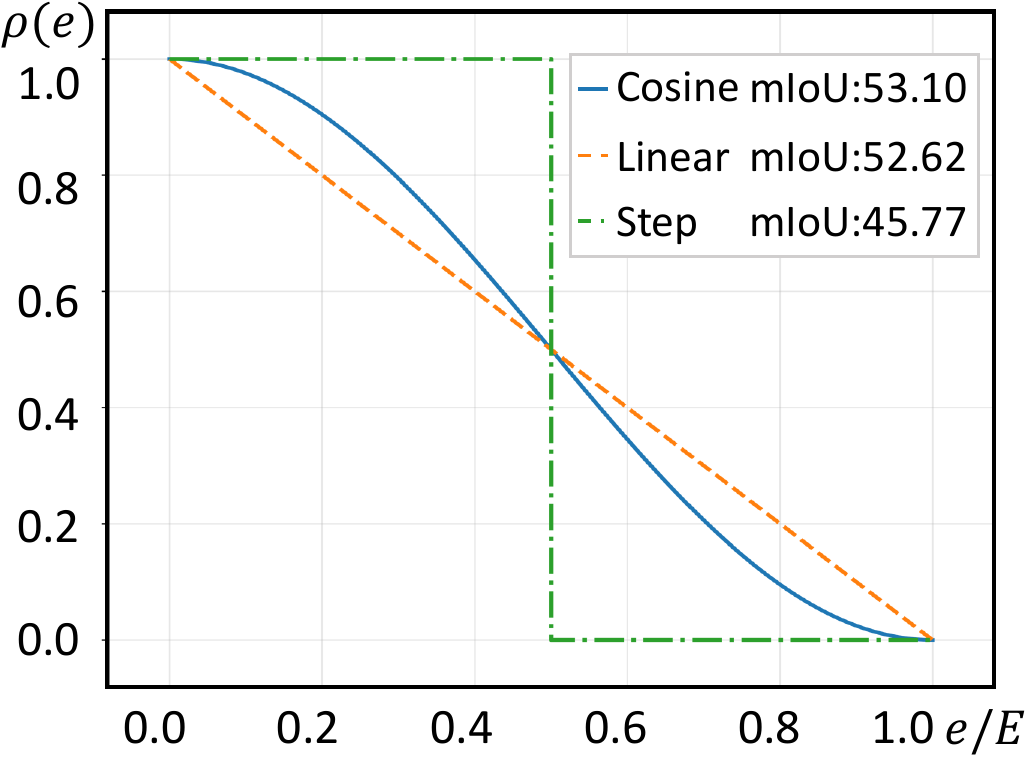}
    \caption{PHC schedules ablation.}
    \label{Fig:ablation_PHC_curve}
\end{minipage}
\hfill
\begin{minipage}[t]{0.6\linewidth}
    \vspace{0pt}
    \centering
    \includegraphics[width=\linewidth]{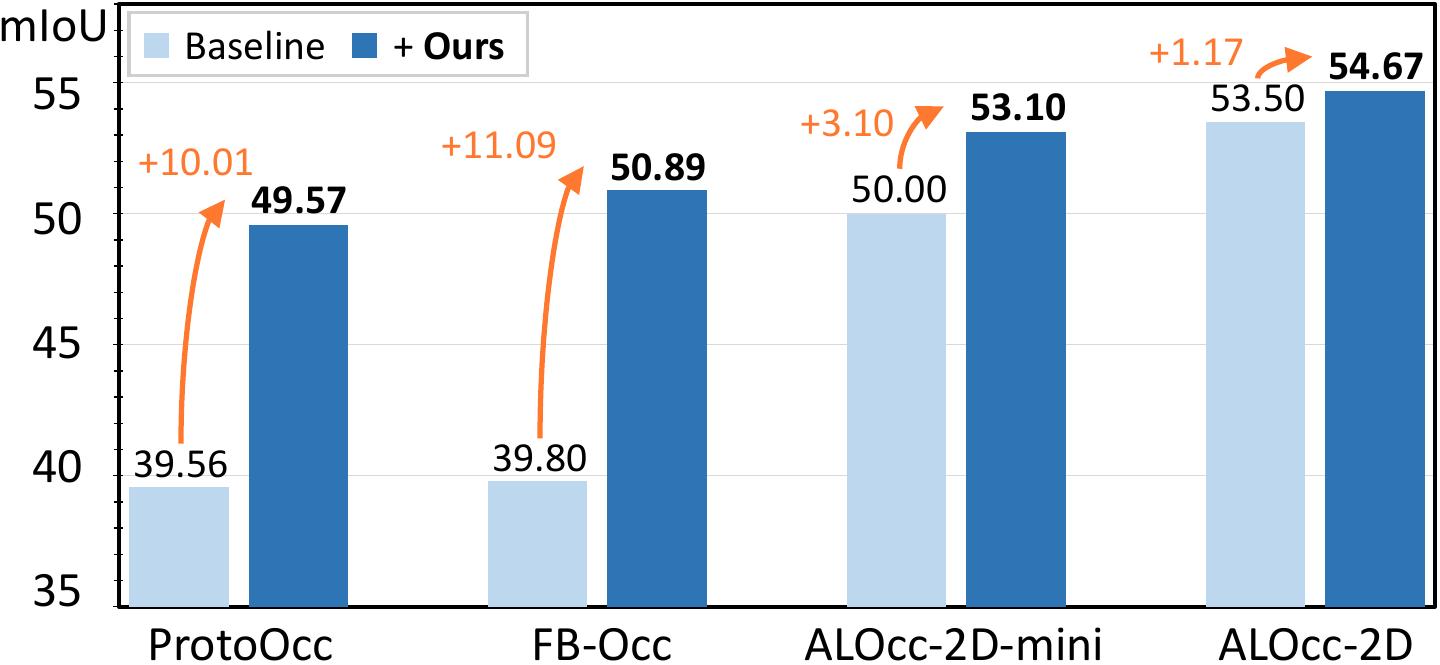}
    \caption{Generalization on different architectures.}
    \label{Fig:generalization}
\end{minipage}
\vspace{-5pt}
\end{figure}

\noindent\textbf{Conditioning Strategy in PHC.}
As shown in Tab.~\ref{tab:cond}, we compare different conditioning strategies in PHC. Using only LiDAR-derived heights as the baseline achieves 52.81 mIoU. In contrast, using ground-truth heights throughout training severely degrades performance to 31.83 mIoU, revealing a large gap between the clean training signal and noisy LiDAR at inference. We further compare two mixing strategies. Linear interpolation blends ground-truth and LiDAR-derived height maps but brings only limited improvement, while our hard replacement strategy achieves the best performance of 53.10 mIoU. Moreover, we ablate different PHC schedules in Fig.~\ref{Fig:ablation_PHC_curve}. The step schedule causes an abrupt transition and degrades performance by 7.33 mIoU, whereas the cosine schedule gradually shifts from ground-truth to LiDAR-derived heights and achieves the best result.

\vspace{-8pt}


\subsection{Generalization Capability}
We further incorporate our design into different occupancy frameworks to evaluate its generalization capability, including ProtoOcc~\cite{kim2025protoocc}, FB-Occ~\cite{li2023fb}, and the ALOcc-2D series~\cite{chen2025alocc}. For ProtoOcc and ALOcc, we insert the proposed HGR before the occupancy head, while for FB-Occ, we directly replace the standard backward projection with HGR. All models are trained with PHC. As shown in Fig.~\ref{Fig:generalization}, our method consistently improves the baseline models across different architectures. Specifically, applying our method to the camera-based ProtoOcc and FB-Occ brings substantial gains of 10.01 and 11.09 mIoU, respectively. When applied to ALOcc variants that utilize LiDAR depth ground truth for feature lifting, our method still achieves improvements of 3.10 and 1.17 mIoU. These consistent gains demonstrate that the proposed HGR and PHC are not tied to a specific framework, but can serve as general and effective plug-and-play modules for 3D occupancy prediction pipelines.

\section{Conclusion}
\label{Sec:Con}
In this paper, we present HiPR, a novel camera--LiDAR framework that reparameterizes the projection space using the LiDAR height prior. By encoding LiDAR observations into a BEV height map, HiPR introduces height-guided sampling and height-validity mask to adaptively adjust sampling range and grid validity, thereby improving geometric alignment in the projection space. In addition, we leverage ground-truth heights during training to stabilize optimization. Extensive experiments demonstrate the superiority of HiPR in accuracy, efficiency, and generalization.

\noindent\textbf{Limitation.}
Although HiPR achieves strong performance, it relies on LiDAR-derived height prior, which can become sparse and less reliable in distant regions. In addition, modeling scene geometry with a single height representation may be insufficient for complex multi-layer structures. We leave the exploration of richer geometric representations and more robust height prior as future work.

\noindent\textbf{Broader Impact.}
With the rapid development of multi-modal autonomous driving systems, improving the accuracy and efficiency of 3D scene understanding has become increasingly important. HiPR represents a step forward in this direction by enhancing geometric alignment in occupancy prediction while maintaining real-time performance. Given the broad relevance of 3D scene understanding, our approach has the potential to benefit a wide range of autonomous driving applications.

\newpage
{   
    \bibliographystyle{ieeenat_fullname}
    \bibliography{ref}
}

\newpage


\appendix

\section{Technical appendices and supplementary material}

\begin{table}[ht]
\centering
\footnotesize
\caption{Comparison of computational cost. Metrics were measured on an NVIDIA 4090 GPU.}
\setlength{\tabcolsep}{10pt} 
\begin{tabular}{lccccc}
\hline
\multicolumn{1}{l}{Method} & mIoU & FPS & Parameters (M) & Training Memory (GB) & GPU Hours\\ 
\hline
FB-Occ \cite{li2023fb} & 39.80 & 14.4 & 68.53 & 6.01 & 210 \\
+\textbf{Ours}  & 50.89 & 14.8 & 68.53 & 5.97 & 212 \\
\hline
ALOcc-2D-mini \cite{chen2025alocc}  & 50.00 & 30.5 & 36.23 & 2.15 & 62 \\
+\textbf{Ours} & 53.10 & 20.8 & 37.21 & 2.58 & 68 \\
\hline
ALOcc-2D \cite{chen2025alocc}  & 53.50 & 8.1 & 63.70 & 3.38 & 122 \\
+\textbf{Ours}  & 54.67 & 7.5 & 63.96 & 4.58 & 134 \\

\hline
\end{tabular}
\label{Tab:computation_comparision}
\end{table}

\begin{figure}[h]
    \centering
    \begin{minipage}[t]{0.48\columnwidth}
        \centering
        \includegraphics[width=\linewidth]{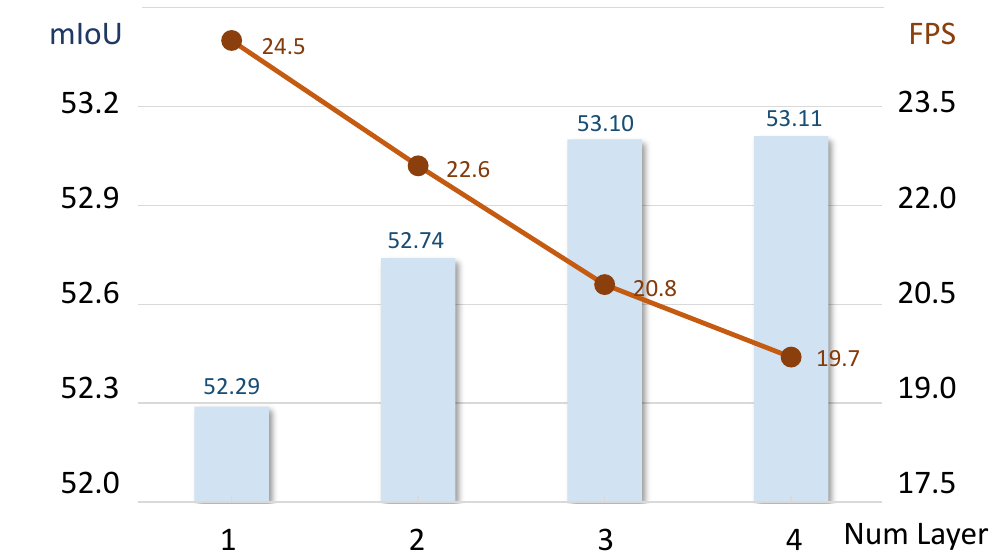}
        \caption{Ablation on HGP layers.}
        \label{Fig:layer_fps}
    \end{minipage}
    \hfill
    \begin{minipage}[t]{0.48\columnwidth}
        \centering
        \includegraphics[width=\linewidth]{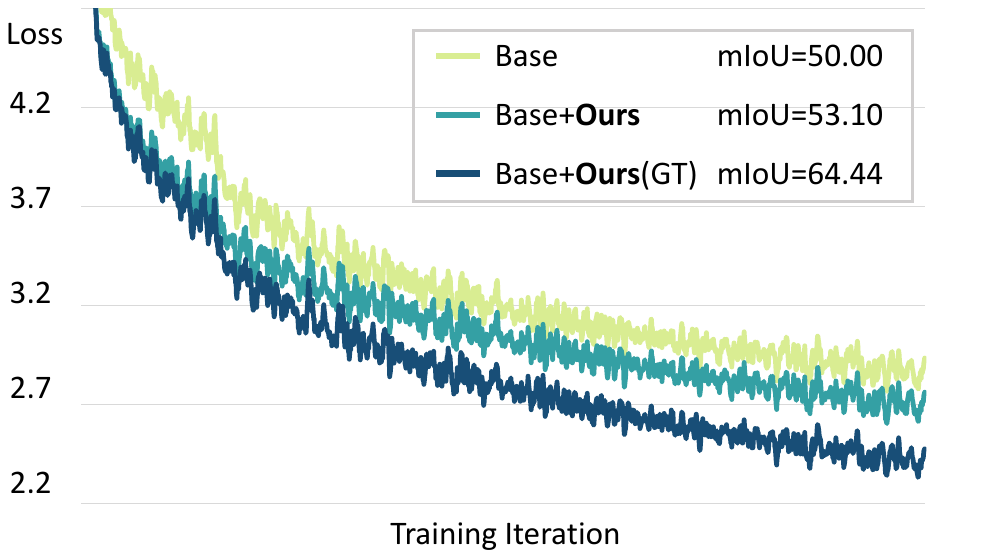}
        \caption{Upper-Bound analysis.}
        \label{Fig:height_upper_bound}
    \end{minipage}
\end{figure}
\vspace{-5pt}

\subsection{Ablation on HGP Layers}
We conduct this ablation by varying the number of HGP layers in HiPR-mini. As shown in Fig.~\ref{Fig:layer_fps}, mIoU performance consistently improves as the number of layers increases, but the gain becomes marginal beyond three layers, while FPS continues to decrease with additional computation. Therefore, we adopt three HGP layers to achieve a better accuracy-efficiency trade-off.

\subsection{Computational Cost}
Tab.~\ref{Tab:computation_comparision} presents the computational cost analysis of our method on different models. Overall, our method consistently improves prediction accuracy with only limited additional computational overhead. Specifically, on FB-Occ~\cite{li2023fb}, our method achieves a 27.9\% improvement in mIoU. Meanwhile, by filtering out BEV grids with the height validity mask, our method reduces unnecessary computation and maintains comparable efficiency. For the LiDAR-depth-assisted ALOcc series~\cite{chen2025alocc}, our method also brings consistent performance gains. On ALOcc-2D-mini, the mIoU increases from 50.00 to 53.10, with only a moderate increase in parameters and training memory. On ALOcc-2D, our method further improves the mIoU from 53.50 to 54.67, while maintaining comparable inference speed. These results demonstrate that our method can effectively enhance occupancy prediction accuracy across different baseline models without introducing prohibitive computational costs.

\subsection{Upper-Bound Analysis}
To evaluate the potential of height guidance, we compare the baseline ALOcc-2D-mini~\cite{chen2025alocc}, our LiDAR-based height-guided variant, and an upper-bound model that leverages ground-truth height maps. Notably, the upper-bound setting uses ground-truth height not only during training but also at inference time. As illustrated in Fig.~\ref{Fig:height_upper_bound}, incorporating LiDAR height guidance improves the baseline by 3.10 mIoU, demonstrating the effectiveness of our reparameterization strategy. Furthermore, ground-truth heights guidance accelerates convergence and yields substantially better performance, improving mIoU by 14.44. These results suggest that a more accurate height prior provide stronger geometric constraints and highlight the upper-bound potential of our height-guided design.

\subsection{Additional Visual Results}
\begin{figure}[t]
    \centering
    \includegraphics[width=1.0\columnwidth]{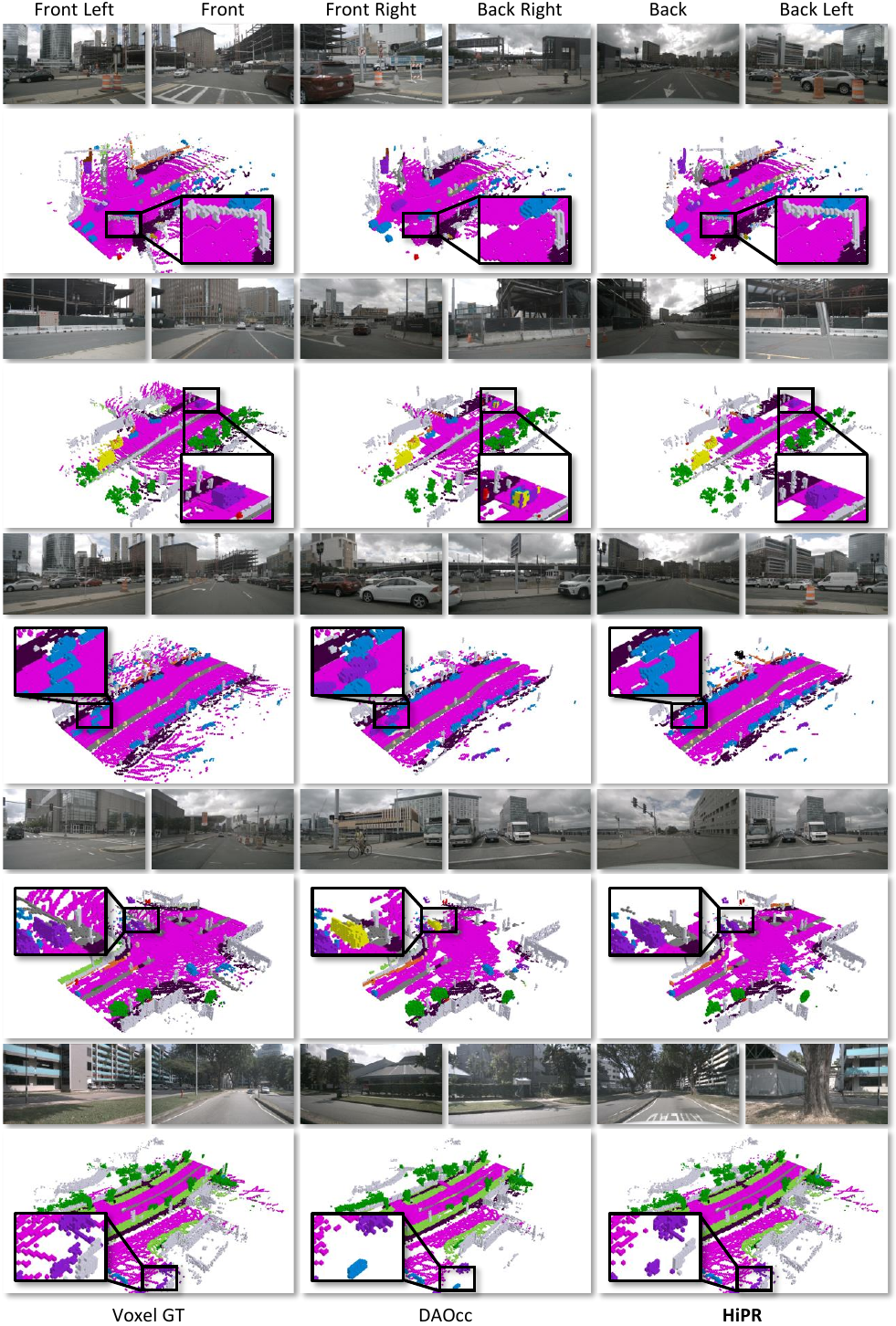}
    \caption{Qualitative comparison between DAOcc and our HiPR on the Occ3D validation set.}
    \label{Fig:occ_vis_supp}
\end{figure}

As shown in Fig.~\ref{Fig:occ_vis_supp}, we provide additional visualizations of DAOcc~\cite{yang2025daocc} and our HiPR on Occ3D.

\clearpage


\newpage

\end{document}